%% file: main.tex
\documentclass{article}
\usepackage{iclr2027_conference,times}

\input{imports}
\input{macros}

\title{Does Random Search Discover Task Experts?\\
Revisiting Neural Thickets}

\author{Julian Asilis\thanks{Equal contribution.} \\
University of Southern California \\
\texttt{asilis@usc.edu}
\And
Guanzhe Hong\footnotemark[1] \\
University of Oxford \\
\texttt{guanzhe.hong@eng.ox.ac.uk}
\AND
Vatsal Sharan \\
University of Southern California \\
\texttt{vsharan@usc.edu}
}

\iclrfinalcopy
\begin{document}

\maketitle
\lhead{}
\vspace{-20 pt}

\input{sections/abstract}

\vspace{-20pt}
\vfill
\input{figures/headline}
\input{sections/introduction}
\input{sections/methods}
\input{sections/results}
\input{sections/technical_details}
\input{sections/discussion}

\input{statements}

\bibliography{refs}
\bibliographystyle{iclr2027_conference}

\newpage 
\appendix
\input{appendix}

\end{document}

%% file: imports.tex
\usepackage{amsmath,amssymb,amsthm}
\usepackage{mathtools}
\usepackage{nicefrac}
\usepackage{xfrac}
\usepackage{bbm,bm}

\usepackage{graphicx}
\usepackage{booktabs}
\usepackage{array}
\usepackage{wrapfig}
\usepackage{float}
\usepackage{afterpage}

\usepackage{tikz}
\usetikzlibrary{arrows.meta,calc,fit,positioning}
\usepackage{pgfplots}
\pgfplotsset{compat=newest}

\usepackage{algorithm}
\usepackage{algpseudocode}

\usepackage{xcolor}
\usepackage{enumitem}
\usepackage{xparse,expl3}
\usepackage{microtype}
\usepackage{breakcites}

\definecolor{lightblue}{RGB}{0,102,204}
\usepackage[
    colorlinks=true,
    linkcolor=black,
    citecolor=black,
    urlcolor=black
]{hyperref}

\usepackage{aliascnt}
\usepackage[capitalize]{cleveref}

\makeatletter
\let\c@table\c@figure
\makeatother

\newaliascnt{proposition}{theorem}

\aliascntresetthe{proposition}

\newaliascnt{lemma}{theorem}

\aliascntresetthe{lemma}

\newaliascnt{corollary}{theorem}

\aliascntresetthe{corollary}

\newaliascnt{definition}{theorem}

\aliascntresetthe{definition}

\newaliascnt{example}{theorem}

\aliascntresetthe{example}

\newaliascnt{remark}{theorem}

\aliascntresetthe{remark}

\newaliascnt{conjecture}{theorem}

\aliascntresetthe{conjecture}

\newaliascnt{observation}{theorem}

\aliascntresetthe{observation}

\crefname{theorem}{Theorem}{Theorems}
\Crefname{theorem}{Theorem}{Theorems}
\crefname{proposition}{Proposition}{Propositions}
\Crefname{proposition}{Proposition}{Propositions}
\crefname{lemma}{Lemma}{Lemmas}
\Crefname{lemma}{Lemma}{Lemmas}
\crefname{corollary}{Corollary}{Corollaries}
\Crefname{corollary}{Corollary}{Corollaries}
\crefname{definition}{Definition}{Definitions}
\Crefname{definition}{Definition}{Definitions}
\crefname{example}{Example}{Examples}
\Crefname{example}{Example}{Examples}
\crefname{remark}{Remark}{Remarks}
\Crefname{remark}{Remark}{Remarks}
\crefname{conjecture}{Conjecture}{Conjectures}
\Crefname{conjecture}{Conjecture}{Conjectures}
\crefname{observation}{Observation}{Observations}
\Crefname{observation}{Observation}{Observations}
\crefname{problem}{Open Problem}{Open Problems}
\Crefname{problem}{Open Problem}{Open Problems}
\crefname{equation}{Equation}{Equations}
\Crefname{equation}{Equation}{Equations}
\crefname{section}{Section}{Sections}
\Crefname{section}{Section}{Sections}
\crefname{subsection}{Section}{Sections}
\Crefname{subsection}{Section}{Sections}
\crefname{appendix}{Appendix}{Appendices}
\Crefname{appendix}{Appendix}{Appendices}

%% file: macros.tex
\newcommand{\nc}{\newcommand}
\newcommand{\rnc}{\renewcommand}

\rnc{\epsilon}{\varepsilon}

\nc{\R}{\mathbb R}
\nc{\Q}{\mathbb Q}
\nc{\Z}{\mathbb Z}
\nc{\N}{\mathbb N}

\let\E\relax
\DeclareMathOperator*{\E}{\mathbb{E}}

\DeclareMathOperator{\Var}{Var}

\nc{\CA}{\mathcal A}
\nc{\CC}{\mathcal C}
\nc{\CD}{\mathcal D}
\nc{\CF}{\mathcal F}
\nc{\CG}{\mathcal G}
\nc{\CH}{\mathcal H}
\nc{\CI}{\mathcal I}
\nc{\CP}{\mathcal P}
\nc{\CR}{\mathcal R}
\nc{\CS}{\mathcal S}
\nc{\CX}{\mathcal X}
\nc{\CY}{\mathcal Y}
\nc{\CV}{\mathcal V}

\nc{\defn}[1]{\textbf{#1}}
\nc{\edefn}[1]{\emph{\textbf{#1}}}
\nc{\Unif}{\mathrm{Unif}}
\nc{\one}{\mathbf{1}}
\nc{\ol}[1]{\overline{#1}}
\nc{\wt}[1]{\widetilde{#1}}
\nc{\restr}[2]{{#1}\!\upharpoonright_{#2}}
\nc{\eps}{\varepsilon}

\nc{\randopt}{\textsc{RandOpt}}
\nc{\ttmv}{\textsc{TT-MV}}
\nc{\anchor}{\textsc{Anchor}}
\nc{\pp}{\,\mathrm{pp}}
\nc{\indep}{\perp\!\!\!\perp}

%% file: sections/abstract.tex
\begin{abstract}
\vspace{-5 pt}

Pretrained models are typically viewed as generalists, requiring bespoke post-training to achieve
task-specific expertise. \citet{gan2026neuralthickets} recently proposed a striking alternative: that
pretrained models lie in a dense ``neural thicket'' of task specialists, which can be surfaced with simple Gaussian weight perturbations.
We examine this neural thicket of task-specific experts and
identify several key findings which question this perspective.
First, we find that the apparent lift of the best-of-$N$ selected perturbation
decays quickly with the size $m$ of the selection set,
strongly suggesting overfitting to the selection questions.
In fact, for
$10$ of the $12$ model--task pairs we study, a simple null model of pure selection noise predicts
more than $75\%$ of the top expert's reward lift.
Second, the top experts' advantage often vanishes on held-out questions, or remains within the
run-to-run variability of the unperturbed pretrained model.
Third, we find that majority voting over the top-$K$ selected perturbations---coined RandOpt($N, K$) by
\citet{gan2026neuralthickets}---is often matched or outperformed by majority voting over $K$ samples
from the base model.
Overall, our results suggest that random weight perturbations may be ineffective at uncovering task
experts around pretrained models, and that task adaptation
requires more sophistication than random search.
Code and artifacts are available at
{\small\href{https://github.com/asil012/RevisitingNeuralThickets}
{\texttt{github.com/asil012/RevisitingNeuralThickets}}}.
\end{abstract}

%% file: figures/headline.tex
\begin{figure}[H]
    \centering
    \includegraphics[width=\textwidth]{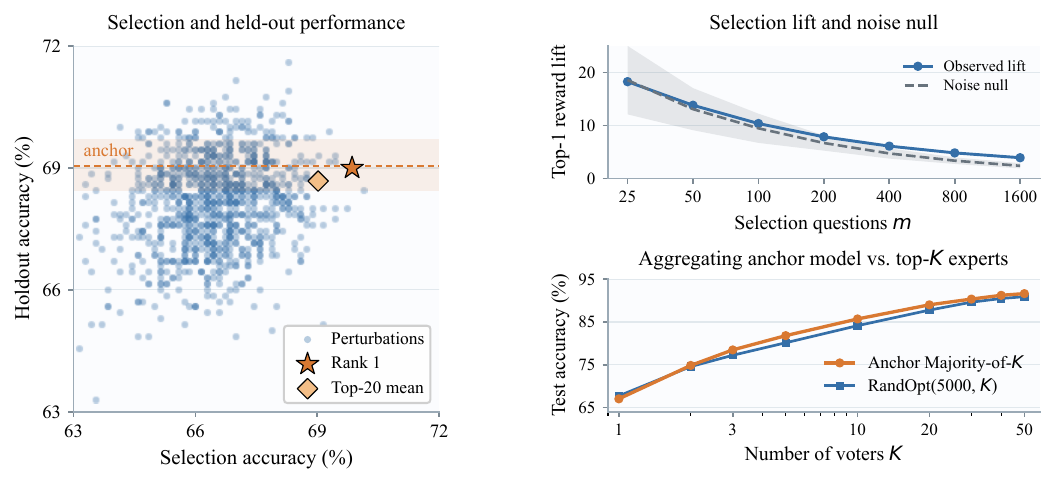}
    \par\vspace{-10pt}
    \small
    \caption[OLMo-3-7B-Instruct on Countdown.]{OLMo-3-7B-Instruct on Countdown. \textbf{Left:}
    Performance of $N=1,000$ perturbations at $\sigma=0.001$; the anchor line/band displays its five-run mean/range on a fixed runtime.
    Note little relation between selection and held-out performance.
    \textbf{Top right:} Reward difference between the best-performing perturbation (``top-1 expert'') at $N=1,000$ and the perturbation population mean on the selection set of size $m$. A null model of pure evaluation noise (shown with $\pm2$ SD) predicts most observed lift, e.g., $85\%$ at $m=200$.
    \textbf{Bottom right:} RandOpt versus anchor majority-of-$K$, at $N = 5,000$ and $\sigma \in \{0.5,1, 2\} \times 10^{-3}$. Anchor voting leads at every $K\ge3$.}
    \label{fig:headline}
\end{figure}

%% file: sections/introduction.tex
\section{Introduction}
\label{sec:intro}

Post-training stands as a sprawling and often expensive component of language-model development.
Techniques such as instruction
tuning \citep{wei2022finetuned}, preference optimization \citep{rafailov2023dpo}, and reinforcement
learning with learned rewards \citep{ouyang2022instructions} all serve to improve a pretrained
model's performance on a particular task,
often with substantial demands on data and compute. Against this
backdrop, \citet{gan2026neuralthickets} proposed a surprising shortcut based on random search.
Their method, RandOpt$(N, K)$, begins with a pretrained model $\theta$ and a
selection set $S_{\mathrm{sel}}$ of questions for a particular task $\mathcal{T}$.
It then evaluates $N$ Gaussian weight perturbations $\{\theta_i\}_{i \leq N}$
of $\theta$ on $S_{\mathrm{sel}}$, and preserves the top-$K$ performing perturbations.
Nominally,
each perturbation adds independent Gaussian noise to every parameter of the pretrained model:
$\theta_i = \theta + \mathcal{N}(0, \sigma^2)$. The released implementation instead couples
noise across equal-shaped tensors; we analyze this detail in Section~\ref{sec:technical}.
At test
time, RandOpt$(N, K)$ aggregates the $K$ selected models by majority vote.
Each perturbed model $\theta_i$ is then thought of as an attempt at exhibiting a task expert for $\mathcal{T}$; the top such experts ought to compete with a genuinely post-trained version of
$\theta$.

Surprisingly, \citet{gan2026neuralthickets} reported that this simple form of random search can
produce task-specific experts.
On the Countdown arithmetic task, for instance, they report that a single top-selected perturbation
of OLMo-3-7B-Instruct achieves about $74\%$ accuracy, compared with about $65\%$ for the unperturbed
``anchor'' model. At the ensemble level, they report that RandOpt can even outperform reinforcement-learning baselines
on reasoning tasks such as GSM8K and MBPP \citep[Table~4]{gan2026neuralthickets}.
The authors attribute this performance to a \emph{neural thicket}: a dense population of
task-specific experts surrounding the pretrained weights.
In their account, pretraining balances
multiple task objectives, and a model optimized for the aggregate loss may remain close to better
solutions for individual tasks. The neural-thicket hypothesis further proposes that such experts
are sufficiently abundant for random perturbations alone to recover them, without additional
training.

However, several important aspects of the neural thicket merit further study. First, \citet{gan2026neuralthickets} primarily report the performance of individual perturbations (i.e., top experts) on the selection set, often reserving the held-out test set for the final RandOpt ensemble.
This leaves it unclear to what extent the top-expert advantage arises from selection effects, rather than a generalizing advantage that persists on new held-out problems.
A priori, these effects could be substantial for the setup of \citet{gan2026neuralthickets}, which selects among thousands of perturbations using only $m=200$ selection questions.
Furthermore, it remains unclear to what extent RandOpt's performance is driven by the advantage (and/or diversity) of the top-$K$ experts, as opposed to the simple benefits of majority voting over $K$ samples from the anchor model. This motivates us to closely study two questions: \textbf{Q1}: How much of the apparent advantage of top-ranking experts persists on held-out questions, as opposed to arising from selection effects? \textbf{Q2}: How much does majority voting over the top-$K$ experts improve performance over majority voting over $K$ samples from the anchor model?

We begin our study of Q1 by testing a null hypothesis of pure selection noise. In this model, each perturbed model has the same expected performance as the anchor, but suffers \emph{independent} noise when evaluated at each question.
(Intuitively, the anchor model is a random variable, and each perturbed model is one crystallization of that random variable.)\footnote{One could also think of the anchor model as being stochastic, and of each perturbed model as a single rollout. Our primary experiments use greedy decoding, however, in keeping with \citet{gan2026neuralthickets}.} A crucial feature of this model---in which there is no neural thicket of task experts---is that the top expert's performance on the selection set $S_{\mathrm{sel}}$ should decay quickly as a function of its size, $m = |S_{\mathrm{sel}}|$. Intuitively, a larger selection set reduces the variance of each expert's performance, lowering the chances of observing a very lucky expert. In contrast, were there to exist an expert with a persistent performance advantage, its expected lift over the perturbation population would remain stable as $m$ grows.

We find that the null model indeed predicts a large majority of the top experts' apparent advantage on the selection set, and that this advantage decays quickly in $m$ (\cref{fig:headline}, top right).
For instance, at $m=200$, among 1,000 OLMo-3-7B-Instruct perturbations on Countdown, the null predicts $6.72$ of the $7.89$ reward points gained over the perturbation average---approximately $85\%$.
Furthermore, the decline in the top expert's apparent advantage (on the selection set) continues well beyond the $m=200$ questions used in \citet{gan2026neuralthickets}.
On Countdown, OLMo sees its top expert advantage drop from $7.89$ points at $m=200$ to
$3.94$ at $m=1600$. This strongly suggests that at least half of the top expert's apparent advantage is due to selection effects.

We then examine performance on held-out questions. For OLMo-3-7B-Instruct across the Countdown, GSM8K, and MBPP reasoning tasks, we find that the rank-1 experts' advantage nearly vanishes on held-out data, with a mean gain of only $0.16$
accuracy points across the three tasks.
Notably, this lands within the run-to-run variability of the anchor model on the same held-out questions (even with greedy decoding!), suggesting that top experts differ little from the anchor.
On Countdown, the $0.44$-point gain over the anchor from our full $N = 5,000$ candidate
search falls far short of the single-expert improvement reported by Gan \& Isola. We discuss this
reproduction gap and our runtime configuration in Appendices~\ref{app:protocol}
and~\ref{app:single}.

We also examine the Qwen2.5-Instruct family of models at the 3B, 7B, and 32B scales. We find that their experts struggle similarly on GSM8K and MBPP, with held-out gains of only about $1\%$. On Countdown, however, Qwen-3B and 7B retain considerable gains of $12.2$ and $11.4$ accuracy points, respectively, over their anchors.
Notably, about 83--94\% of these gains are driven by improved validity of generated expressions, rather than improved arithmetic ability. That is, top experts produce valid expressions more often than the anchor, which increases accuracy even if the conditional correctness of valid expressions remains similar.
Overall, for 10 of the 12 model--task settings we consider, our results strongly suggest that the apparent advantage of top experts is largely due to selection effects, and that their held-out performance often lies within the run-to-run variability of the anchor model.

We then turn to Q2, and compare RandOpt's majority-vote aggregation of the top-$K$ experts with simple test-time majority voting (TT-MV) over $K$ samples from the anchor model.
At $K=50$, across OLMo-3-7B-Instruct and Qwen2.5-Instruct at 3B, 7B, and 32B on
$\{$Countdown, GSM8K, MBPP$\}$, we find that RandOpt is statistically indistinguishable from TT-MV
in 10 of the 12 model--task settings. In the remaining 2 settings---Qwen2.5-Instruct at 7B and 32B
on Countdown---RandOpt slightly outperforms TT-MV by about $2\%$ (\cref{tab:voting-main}). This suggests that in most cases, the benefits of RandOpt's majority-vote aggregation are largely matched by the benefits of majority voting over the anchor model.
Overall, our results question the existence of a dense neural thicket, and suggest that eliciting
genuine task expertise from pretrained models requires more sophistication than random search.

%% file: sections/methods.tex
\section{Preliminaries and Methodology}
\label{sec:methods}

\subsection{The neural thicket}

Let us briefly review the neural thicket findings reported by \citet{gan2026neuralthickets}. We use
their language throughout the paper, referring to pretrained models as ``anchors,'' and to
top-performing weight perturbations of pretrained models as ``experts.''

\textbf{Claim 1: Task experts are dense in weight space.} The first and most central claim of
\citet{gan2026neuralthickets} is that pretrained models are closely surrounded in weight space
by task experts. This effect is sufficiently pronounced that simple Gaussian weight perturbations of
the pretrained model can surface task experts with shockingly high probability (e.g., $>10\%$).
In fact, on some model--task pairs, as many as 45\% of random weight perturbations
produce a task expert that outperforms the anchor by 15+\%! (See their Figure~3(a).)
This suggests, strikingly, that pretraining produces not just the weight vector of a single
performant model, but an entire weight-space neighborhood hosting a ``veritable thicket of
task-specific solutions'' \citep[p.~1]{gan2026neuralthickets}.

\textbf{Claim 2: RandOpt competes with post-training.}
Perhaps the second major claim of Gan \& Isola is that simple aggregations of such experts
can be used to match or even outperform standard post-training methods.
For instance, for OLMo-3-7B-Instruct on the Countdown reasoning task, it is reported that an
ensemble of the top 50 experts---i.e., RandOpt with $K=50$---outperforms ensembles of
GRPO post-trained models by 5 points (their Figure 1(c)).

\textbf{Claim 3: Expert density increases with model size.}
Finally, Gan \& Isola report that the density of task experts increases with model size.
Roughly speaking, the neural thicket can sometimes be difficult to observe for models at the 0.5B
or 1.5B scales, but appears fairly consistently at the 3B+ and 7B+ scales. In fact, when measuring
the proportion of random perturbations that outperform the anchor, it can appear to increase
consistently with model size (see their Figures 3(a) and 8). This points to an intriguing scaling
axis of the neural thicket effect that merits further study.

\subsection{RandOpt and the evaluation setting}

We study three reasoning tasks used by Gan \& Isola: Countdown, GSM8K, and MBPP.
Countdown asks for an expression reaching a target integer using each of
three or four supplied
integers exactly once and only operations from $\{+,-,\times,\div\}$.
GSM8K \citep{cobbe2021gsm8k} comprises grade-school word problems requiring two to eight
arithmetic steps and a numerical answer. MBPP \citep{austin2021program} asks for Python
programs satisfying natural-language specifications and three tests.

The pretrained models in our main searches are
OLMo-3-7B-Instruct and Qwen2.5-Instruct at the 3B, 7B, and 32B sizes
\citep{olmo2025olmo3,qwen2025qwen25}.
Recall that RandOpt$(N, K)$ samples $N$ candidates
around an anchor with weights $\theta$,
\begin{equation}
    \theta_i=\theta+\sigma_i \, \epsilon_i,\qquad i\in[N],
    \label{eq:candidates}
\end{equation}
where $\sigma_i$ controls the perturbation size and $\epsilon_i$ is
nominally standard
Gaussian noise with independent entries. In the released implementation we evaluate, equal-shaped
tensors receive coupled noise; see Appendix~\ref{app:implementation}. RandOpt$(N, K)$ then ranks all candidates by their mean
reward on $m$ selection questions and retains the top $K$.
At test time, it aggregates these
top-$K$ models via majority vote over valid extracted answers.
We use the released RandOpt implementation, which samples each
$\sigma_i \in \{0.5,1,2\} \times 10^{-3}$ uniformly at random.
For selection, Countdown uses answer correctness plus $0.1$ times a format score, while
GSM8K and MBPP use binary task rewards; see Appendix~\ref{app:protocol}.

For OLMo and Qwen-3B/7B, we evaluate $N=5,000$ candidates on $m=200$ selection questions;
for Qwen-32B, we use $N=1,000$. We then evaluate the top 50 experts on 1,400 held-out Countdown
questions, 1,319 GSM8K questions, and 500 MBPP tasks.
Following
\citet{gan2026neuralthickets}, candidate and single-model evaluations use greedy decoding. Anchor
TT-MV samples at temperature $0.6$ and top-$p=0.95$.

\subsection{Measuring selection and transfer}
\label{sec:null}

We first examine a statistical model of expert performance under the null hypothesis that
candidates have no persistent differences in expected reward.
Selecting among $N=5{,}000$
candidates can then yield a substantial apparent advantage through finite-sample variation
alone. In particular, under the \emph{selection-noise null}, the reward $Y_{ij}$ of perturbation $i$ on question $j$
follows
\begin{equation}
    Y_{ij}=\mu_j+\varepsilon_{ij},\qquad
    \E[\varepsilon_{ij}]=0,\quad \Var(\varepsilon_{ij})=v_j.
    \label{eq:null-reward}
\end{equation}
Here $\mu_j$ is the common expected reward on question $j$ and the errors $\varepsilon_{ij}$ are independent
across candidate--question pairs.
Letting $S_i(m)$ denote candidate $i$'s mean reward on $m$ selection questions, we define the
top expert's performance lift above the perturbation-population mean as
\begin{equation}
    \Delta_{\mathrm{sel}}(m)
      =\max_i S_i(m)-\frac{1}{N}\sum_{i=1}^N S_i(m).
    \label{eq:selection-lift}
\end{equation}
We implement the empirical null by sampling each $Y_{ij}$
with replacement from the observed candidate rewards on question $j$. Averaging over a fixed
set of $m$ selection questions gives $\Var(S_i(m))=v/m$, where $v$ is the mean of the
corresponding per-question variances $v_j$. Approximating the candidate means as Gaussian\footnote{Assuming independent, bounded question-level rewards, the central limit theorem implies that candidate means are asymptotically Gaussian as $m\to\infty$. See Appendix~\ref{app:selection} for further detail on the null model.} then gives
\begin{equation}
    \E[\Delta_{\mathrm{sel}}(m)]\approx a_N\sqrt{v/m},
    \qquad
    a_N=\E\!\left[\max_{i\le N} Z_i\right],\quad Z_i\sim\mathcal N(0,1).
    \label{eq:extreme-value}
\end{equation}
Here $a_N$ is the expected maximum
of $N$ independent standard normal variables---equivalently, the expected $N$th order statistic
of those samples; for example,
$a_{1000}\approx3.241$ and $a_{5000}\approx3.678$.
A key feature of the null model is that for
fixed $N$, the expected lift $\E[\Delta_{\mathrm{sel}}(m)]$ vanishes as $m \to \infty$.
(Intuitively, every perturbed model's mean reward simply converges to the shared population mean.)
In contrast, if there exists any candidate with a persistent reward advantage of
$\epsilon > 0$, then the limiting expected lift is at least $\epsilon$.

%% file: sections/results.tex
\section{Selection, Transfer, and Test-Time Voting}
\label{sec:results}

\subsection{Understanding selection lift and held-out transfer}
\label{sec:selection}

\begin{figure}[t]
    \centering
    \includegraphics[width=\textwidth]{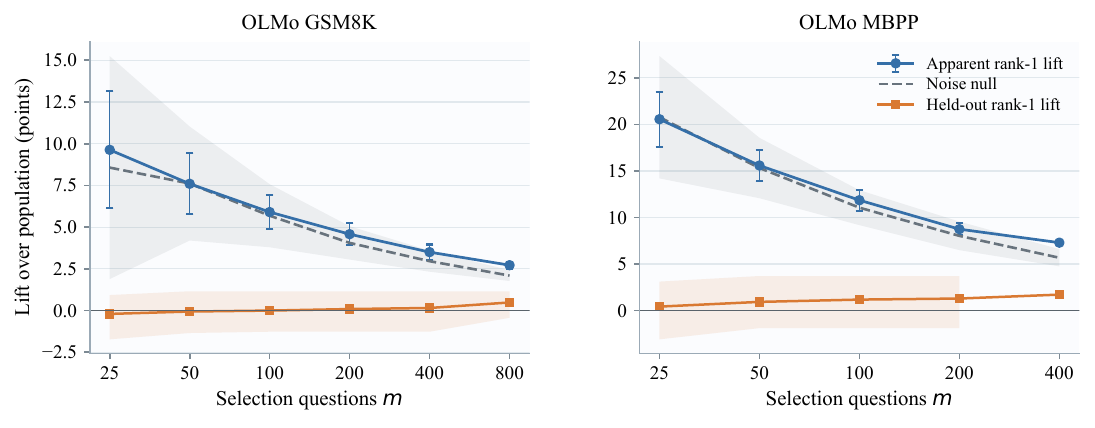}
    \caption{Selection-size sweeps for OLMo-3-7B-Instruct, $N=5{,}000$. Blue: rank-1 selection reward
        lift over the perturbation population, with $\pm1$ SD across selection subsets. Gray: empirical
        noise null with $\pm2$ simulation SD.
        Orange: held-out rank-1 accuracy gain over the perturbation population.
        \textbf{Left:} GSM8K shows little transfer. \textbf{Right:} MBPP has a positive effect at
        larger $m$ on the official 500-task test. ($m=400$ uses the entire selection
        pool, thus lacks selection-subset bands.)}
    \label{fig:selection-transfer}
\end{figure}

\textbf{Top-1 expert lift is largely predicted by selection noise.}
We begin by comparing the selection-noise null (Section~\ref{sec:null}) with the observed reward lift
of the top-selected, or ``rank-1,'' expert.
We find that on each of Countdown, GSM8K, and MBPP, more than 85\% of OLMo's observed rank-1 lift is predicted by the null model (Figures~\ref{fig:headline} (top right) and~\ref{fig:selection-transfer}).
This holds for $m=200$, as used by \citet{gan2026neuralthickets}.
For Qwen2.5 at 3B, 7B, and 32B, the null predicts approximately $76$--$86\%$ of the
rank-1 lift on GSM8K and $83$--$100\%$ on MBPP.
Countdown, however, retains larger residuals at 3B and
7B: the null predicts only $31\%$ and $53\%$ of their observed lifts, respectively
(Appendix Table~\ref{tab:selection-robustness}).

\textbf{OLMo expert lift decays with selection-set size.} We next examine how the rank-1 expert's apparent advantage varies as a function of the selection-set size $m$. A persistent expert advantage would be expected to remain stable as $m$ increases, whereas an advantage decaying in $m$ would strongly suggest (some) overfitting of the top-1 expert to the selection questions.
We find that the top expert's lift indeed decreases sharply with $m$. For OLMo on Countdown, increasing $m$ from 200 to 1600 halves the rank-1 selection lift, from $7.89$ to $3.94$ reward points; GSM8K and MBPP see similar drops (Figures~\ref{fig:headline}, \ref{fig:selection-transfer}).
Furthermore, OLMo's rank-1 experts gain little held-out accuracy over the perturbation population
on GSM8K and MBPP (Figure~\ref{fig:selection-transfer}).

\textbf{Small Qwen models display a persistent advantage.} In contrast to OLMo, we find that Qwen-3B and 7B retain their top experts' advantage on Countdown as $m$ increases. Respectively, their rank-1 lifts decrease only slightly from $24.1$ and $18.2$ points at $m=200$ to $23.8$ and $16.1$ at $m=800$.
This suggests the existence of a persistent advantage enjoyed by their top experts. We investigate the source of this advantage further at the end of Section~\ref{sec:single}.
On GSM8K and MBPP, however, Qwen-3B, 7B, and 32B all see their top experts' advantage decay with $m$, suggesting that the apparent advantage includes selection effects (Appendix Table~\ref{tab:qwen-selection-decay}).

\subsection{Comparing experts with the anchor}
\label{sec:single}

\begin{figure}[t]
    \centering
    \includegraphics[width=\textwidth]{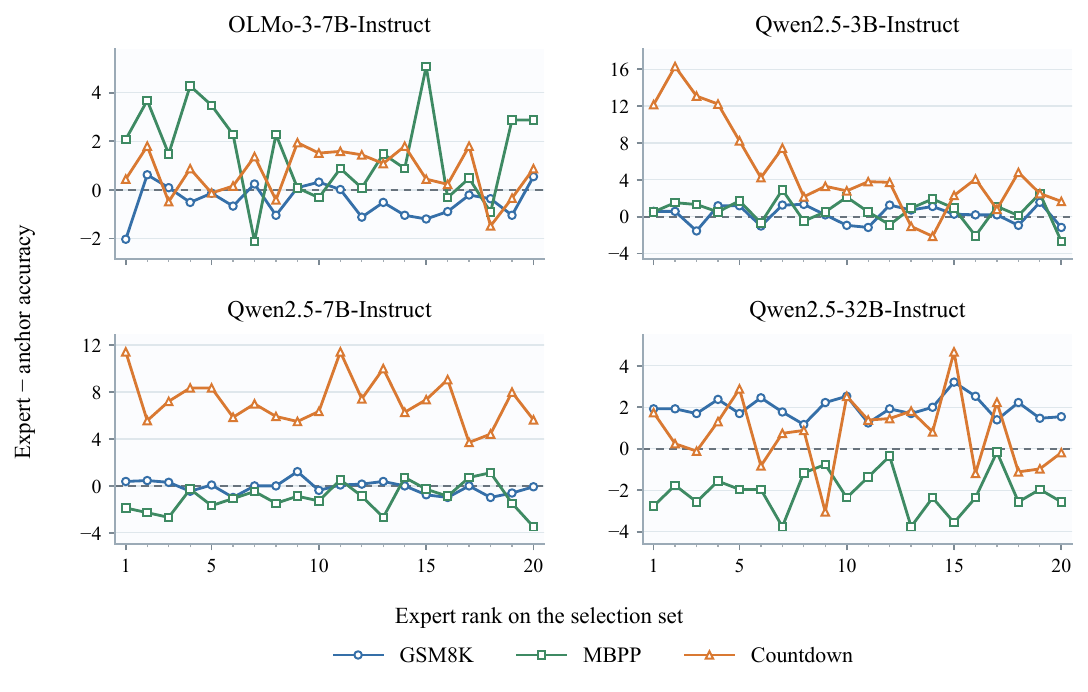}
    \caption{Held-out performance of the top-20 selected perturbations, expressed as percentage point accuracy difference from the anchor.
    Anchor baselines are averaged over five runs per model--task.
    All searches use $N=5,000$ candidates, with the exception of Qwen-32B, which uses $N=1,000$.
    GSM8K and MBPP generally see little advantage from top experts, while Countdown sees substantial
    gains for Qwen-3B and 7B, but not for OLMo or Qwen-32B.
    }
    \label{fig:single}
\end{figure}

\textbf{OLMo's top experts see mixed performance.}
We now compare selected experts with their anchor models on held-out data (Figure~\ref{fig:single}).
For OLMo, we see its rank-1 experts change test accuracy by
$+0.44$ points on Countdown, $-2.03$ on
GSM8K, and $+2.08$ on MBPP (Figure~\ref{fig:single}).
Averaging the accuracies across the top 20 experts, we find that OLMo's top experts achieve average gains over the anchor of $+0.73$ points on Countdown,
$-0.44$ on GSM8K, and $+1.53$ on MBPP.
In short, OLMo's top experts see generally mixed performance across tasks; in
Appendix~\ref{app:selection-subsets} we study the sensitivity of these findings to the
choice of selection subset, and find them to be fairly stable.

\textbf{Most Qwen experts differ little from the anchor.}
On the GSM8K held-out set, the rank-1 experts from Qwen-3B, 7B, and 32B change accuracy by
a relatively modest $+0.58$, $+0.38$, and $+1.93$ points, respectively.
On MBPP, meanwhile, we find the rank-1 expert performance to be mixed, with
anchor differences of $+0.52$, $-1.88$, and $-2.76$ points, respectively.
Averaging over the top-20 experts, we find that Qwen-3B, 7B, and 32B change accuracy by
$+0.24$, $-0.11$, and $+1.95$ points on GSM8K,
and by $+0.62$, $-1.03$, and $-2.08$ points
on MBPP, respectively. For these tasks, the Qwen models do not appear to benefit appreciably from
random weight search at the evaluated search sizes.

\textbf{Qwen-3B and 7B see substantial gains on Countdown, with a catch.}
On Countdown, we find that smaller Qwen models can surface experts with considerable held-out
performance gain. In particular, Qwen-3B's rank-1 expert reaches $26.8\%$ accuracy on the held-out
data against its anchor's $14.6\%$---a striking
$12.2$-point gain---and its top-20 experts
average a $5.1$-point gain. Similarly, Qwen-7B's rank-1 expert
improves upon the
anchor by $11.4$ accuracy points.

However, against the original matched anchor execution, we find that \textit{a substantial part of Countdown gains comes from answering more
questions, not solving them better}. Countdown credits a question only when a valid expression is
extracted, so an expert can gain either by solving attempted questions more accurately or by simply
attempting more questions. Splitting held-out questions by which model produces a valid expression
makes this distinction exact. For Qwen-3B, whose anchor emits a valid expression on only 26\% of
questions, nearly all of the rank-1 expert's
$+11.4$-point gain comes from changing valid-expression coverage: it
produces a valid expression on 345
questions where the anchor does not (with 67.0\% correctness, near the anchor's own conditional
rate) and produces no valid expression on 181 questions where the anchor produces one, while on
the 185 questions both models produce valid expressions it wins 28 and
loses 26, essentially 50-50 (McNemar $p = 0.89$). Qwen-7B repeats the pattern with one difference:
roughly 86\% of its net gain likewise comes from changing
valid-expression coverage, but on the 482 questions
both models produce valid expressions it wins 58 and loses 38 ($p = 0.052$), weak evidence of a small genuine
``reasoning-relevant'' improvement. See Appendix~\ref{app:common-valid} for further detail.

\subsubsection[USPTO: exploiting label imbalance]{USPTO: exploiting label imbalance}
\label{sec:chemistry}

We also examine OLMo's neural thicket for the USPTO-50k chemistry classification task,
where one must assign a chemical reaction to one of ten numbered categories \citep{schneider2016uspto50k}.
Crucially, the categories are imbalanced; category 1 alone covers $30\%$ of questions.
In fact, this percentage surpasses OLMo's anchor performance of $27.5\%$ on held-out data,
meaning that the ``predict class 1'' strategy outperforms the anchor!

At first glance, OLMo's top experts appear to achieve considerable gains over the anchor,
with an average of $5.06$ points across the top 20 experts (Table~\ref{tab:uspto-shuffle}).
However, upon randomly reassigning category numbers per question, we find that about 80\% of
the apparent gains disappear. Notably, the anchor's accuracy remains nearly constant under
this shuffling, while the top expert's absolute accuracy drops considerably. This suggests that the
apparent thicket is largely due to the shortcut of exploiting label imbalance, rather than
``genuine'' task expertise. Nevertheless, a modest held-out
advantage of $1$--$3$ accuracy points persists, though
it remains below the accuracy of the ``predict class 1'' strategy.

\begingroup
\begin{table}[H]
    \centering
    \begin{tabular}{lrrrr}
        \toprule
        Answer choices & Top-1 & Top-5 & Top-10 & Top-20 \\
        \midrule
        Original & $+7.79$ & $+5.21$ & $+5.42$ & $+5.06$ \\
        Shuffled & $+3.20$ & $+1.48$ & $+1.22$ & $+1.12$ \\
        \bottomrule
    \end{tabular}
    \caption{OLMo-3-7B-Instruct at $N=5,000$ perturbations on the USPTO-50k dataset. The first row reports top experts' held-out gain over the anchor in accuracy points. The second reports the same for held-out data that has its category numbers shuffled. Roughly $60$--$80\%$ of the top experts' gains vanishes under shuffling, suggesting that experts primarily learn to abuse label imbalance.}
    \label{tab:uspto-shuffle}
\end{table}
\endgroup

\subsection{RandOpt versus anchor test-time voting}
\label{sec:aggregation}

\input{table}

\textbf{OLMo's TT-MV exceeds RandOpt on Countdown.}
In the headline setting of
\citet{gan2026neuralthickets} for OLMo on Countdown, we find that anchor
majority voting outperforms RandOpt for every evaluated $K\ge3$
(Figure~\ref{fig:headline}, bottom right).
At $K=1$ (i.e., selecting the top expert), RandOpt exceeds the anchor by just $0.44$ points,
which lands within the roughly $1.5$-point range of variability for repeated anchor evaluations.
Notably, this falls far short of the approximately $8$-point single-expert gain reported by
\citet{gan2026neuralthickets}, which we were not able to reproduce using their released code.

\textbf{Qwen sees RandOpt gains on Countdown.}
For Qwen-7B and 32B on Countdown, RandOpt exceeds anchor voting at all three budgets
$K\in\{10,20,50\}$, with fairly stable gains of
$1.5$--$2.7$ accuracy points that are significant
at the $5\%$ level (Table~\ref{tab:voting-main}).
In contrast, Qwen-3B exhibits a strong advantage for RandOpt of about
$7$ points at $K=10$,
which narrows to $2.6$ points at $K=20$ before falling to
$-0.8$ points at $K=50$.
This pattern is consistent with Qwen-3B's strong leading experts and rapidly diminishing
gains at lower ranks (\cref{fig:single}).

\textbf{TT-MV matches RandOpt across GSM8K and MBPP.}
Across GSM8K and MBPP, we find that RandOpt never significantly exceeds anchor TT-MV at
$K\in\{10,20,50\}$ (Table~\ref{tab:voting-main}); gaps range from $-2.12$ to $+0.87$
accuracy points. In fact, the only statistically significant difference favors
anchor voting on Qwen-7B GSM8K
at $K=10$, by $0.84$ points. Anchor voting thus provides a strong no-search baseline,
matching or exceeding RandOpt on most model--task pairs.

\subsection{Increasing diversity of the neural thicket}

    We next examine Gan \& Isola's claim that the density of task experts increases with model size.
For a given evaluation metric (e.g., accuracy) and threshold value (e.g., 105\%), the
\emph{density} of a model's neural thicket is the fraction of random perturbations that exceed the
threshold relative to the population mean of perturbations.\footnote{We note that
Gan \& Isola define density relative to the anchor model, rather than the perturbation population.
We find that using the perturbation population mean helps to isolate the behavior of the perturbations themselves.}

\begin{figure}[!t]
    \centering
    \includegraphics[width=\textwidth]{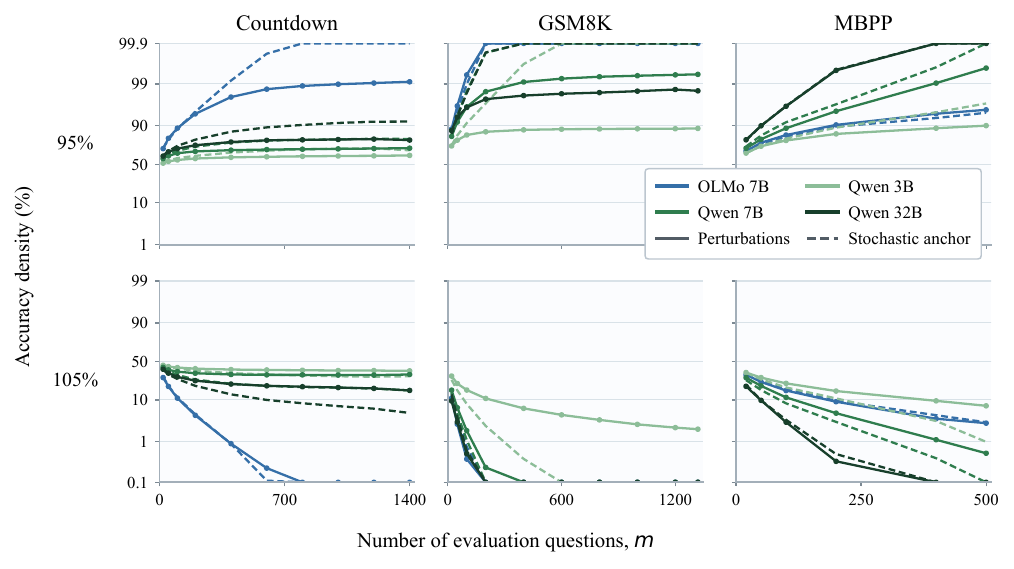}
    \caption{Accuracy density versus evaluation-set size for OLMo
        and Qwen models.
        Each row gives the fraction of accuracies exceeding the indicated percentage of
        their population mean on the same questions. Specifically, the top row uses 95\% and the bottom row 105\%.
        Accuracy values are clipped to $[0.1\%, 99.9\%]$ for display.
        Solid curves use perturbations at $\sigma=10^{-3}$ with greedy decoding; dashed curves
        use 100 rollouts from the anchor model with temperature $0.6$ and top-$p=0.95$. Accuracy
        density often trends toward 0 or 1 with increasing $m$, suggesting that perturbations
        concentrate around their population mean, rather than exhibiting a persistent advantage.
        Stochastic rollouts from the anchor model display qualitatively similar accuracy trends,
        suggesting that the observed behavior may be due to mere sampling noise.
        }
    \label{fig:accuracy-density-m}
\end{figure}

\textbf{Diversity decays with evaluation-set size.} We first observe that for a fixed choice of
threshold $\tau > 100\%$, the density of perturbations exceeding $\tau$ typically decays to 0
with the number of evaluation questions $m$ (Figure~\ref{fig:accuracy-density-m}). Similarly,
for $\tau < 100\%$, the density often tends to 1 with increasing $m$. This behavior appears largely
consistent across models on MBPP and GSM8K, and for OLMo on Countdown. However, the Qwen models would
appear to display persistent behavior on Countdown, with a stable proportion of
favorable and unfavorable perturbations as $m$ increases. This is consistent with our previous
finding that experts from Qwen-3B and 7B can exhibit a persistent advantage on Countdown, while OLMo
and the larger Qwen models do not. Notably, however, we also observe that \emph{qualitatively similar
trends appear even for stochastic rollouts from the anchor model}. This demonstrates that the observed
trends may be an artifact of mere sampling noise, rather than
a persistent advantage among a subpopulation.

\begin{figure}[!t]
    \centering
    \includegraphics[width=\textwidth]{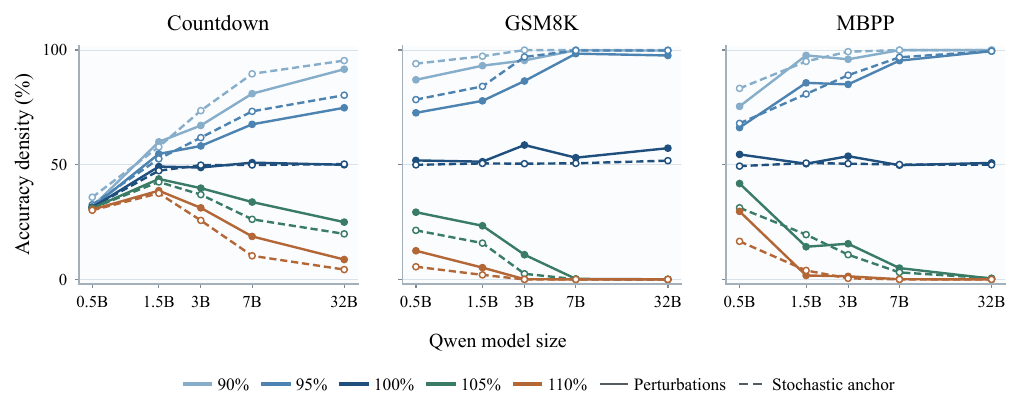}
    \caption{Accuracy density versus Qwen model size on $m=200$ questions. Accuracy density trends toward 0 (respectively 1) for thresholds $\tau > 100\%$ (respectively $\tau < 100\%$), suggesting simple concentration of perturbations around their population mean. Similar trends are observed for simple stochastic rollouts from the anchor model.}
    \label{fig:accuracy-density-size}
\end{figure}

\textbf{Diversity does not appear to increase with model size.} We next examine density as a function of model size
(Figure~\ref{fig:accuracy-density-size}). For thresholds $\tau > 100\%$, we find that the
density of favorable perturbations typically decreases with model size, contrary to Gan \& Isola's
claim. Furthermore, the behavior of stochastic rollouts from the anchor model is again
qualitatively similar to that of the perturbations, suggesting that mere sampling noise may
be responsible for the observed trends.

%% file: table.tex
\begin{table}[t]
\centering
\small
\setlength{\tabcolsep}{4pt}
\definecolor{significanceYes}{HTML}{3E7154}
\definecolor{significanceNo}{HTML}{A15C5A}
\begin{tabular}{llrc@{\hspace{10pt}}rc@{\hspace{10pt}}rc}
\toprule
 & & \multicolumn{2}{c}{$K=10$} & \multicolumn{2}{c}{$K=20$} & \multicolumn{2}{c}{$K=50$} \\
\cmidrule(lr){3-4}\cmidrule(lr){5-6}\cmidrule(lr){7-8}
Model & Task & Difference & Significant
 & Difference & Significant & Difference & Significant \\
\midrule
OLMo 7B & GSM8K & $+0.41$ & \textcolor{significanceNo}{No} & $+0.30$ & \textcolor{significanceNo}{No} & $+0.56$ & \textcolor{significanceNo}{No} \\[1pt]
 & MBPP & $+0.87$ & \textcolor{significanceNo}{No} & $+0.72$ & \textcolor{significanceNo}{No} & $+0.53$ & \textcolor{significanceNo}{No} \\[1pt]
 & Countdown & $-1.57$ & \textcolor{significanceYes}{Yes} & $-1.24$ & \textcolor{significanceYes}{Yes} & $-0.71$ & \textcolor{significanceNo}{No} \\[1pt]
\midrule[0.3pt]
Qwen 3B & GSM8K & $-0.71$ & \textcolor{significanceNo}{No} & $0.00$ & \textcolor{significanceNo}{No} & $-0.61$ & \textcolor{significanceNo}{No} \\[1pt]
 & MBPP & $+0.66$ & \textcolor{significanceNo}{No} & $-0.03$ & \textcolor{significanceNo}{No} & $+0.87$ & \textcolor{significanceNo}{No} \\[1pt]
 & Countdown & $+7.01$ & \textcolor{significanceYes}{Yes} & $+2.57$ & \textcolor{significanceYes}{Yes} & $-0.84$ & \textcolor{significanceNo}{No} \\[1pt]
\midrule[0.3pt]
Qwen 7B & GSM8K & $-0.84$ & \textcolor{significanceYes}{Yes} & $-0.36$ & \textcolor{significanceNo}{No} & $-0.28$ & \textcolor{significanceNo}{No} \\[1pt]
 & MBPP & $-2.12$ & \textcolor{significanceNo}{No} & $-0.87$ & \textcolor{significanceNo}{No} & $-1.94$ & \textcolor{significanceNo}{No} \\[1pt]
 & Countdown & $+2.17$ & \textcolor{significanceYes}{Yes} & $+2.73$ & \textcolor{significanceYes}{Yes} & $+2.25$ & \textcolor{significanceYes}{Yes} \\[1pt]
\midrule[0.3pt]
Qwen 32B & GSM8K & $+0.03$ & \textcolor{significanceNo}{No} & $+0.24$ & \textcolor{significanceNo}{No} & $+0.32$ & \textcolor{significanceNo}{No} \\[1pt]
 & MBPP & $-0.84$ & \textcolor{significanceNo}{No} & $+0.85$ & \textcolor{significanceNo}{No} & $+0.09$ & \textcolor{significanceNo}{No} \\[1pt]
 & Countdown & $+1.50$ & \textcolor{significanceYes}{Yes} & $+2.27$ & \textcolor{significanceYes}{Yes} & $+1.86$ & \textcolor{significanceYes}{Yes} \\[1pt]
\bottomrule
\end{tabular}
\caption{Held-out accuracy of RandOpt minus that of anchor TT-MV, in percentage points at
    the voting budgets $K = \{10, 20, 50\}$. Statistical significance columns record whether a 
    two-sided 95\% paired-question bootstrap interval excludes zero. On GSM8K and MBPP,
    TT-MV matches or outperforms RandOpt. On Countdown, Qwen models 
    prefer RandOpt and OLMo prefers TT-MV.}
\label{tab:voting-main}
\end{table}

%% file: sections/technical_details.tex
\section{Technical Details}
\label{sec:technical}

\textbf{Greedy decoding is nondeterministic.}
Even with greedy decoding, anchor inference is not deterministic. Across five executions of
OLMo-3-7B-Instruct on Countdown in one runtime, final answers disagree by $6.1\%$ on average,
while accuracy spans about one point. This run-to-run variation may obscure a mild persistent
expert advantage (Appendix~\ref{app:fixed-repeats}).

\textbf{Tied weight perturbations and restoration artifacts.}
The released implementation resets the random generator before
each tensor, causing equal-shaped tensors to receive identical noise. It also restores the anchor
by subtracting perturbations in bf16, allowing rounding residuals to accumulate across candidates.
We find that both artifacts have little effect on our results
(Appendix~\ref{app:implementation}).

%% file: sections/discussion.tex
\section{Discussion and Conclusion}
\label{sec:discussion}

We study the neural thicket recently proposed by \citet{gan2026neuralthickets}, which posits that
pretrained models lie in a region of weight space densely populated by task experts.
In most settings we consider, we find that the apparent advantage of top-ranking
experts on the selection set is largely explained by selection effects, and
decays quickly in the size of the selection set. On held-out questions, we find that top-expert performance often differs little from that of the unperturbed anchor model.

Overall,
pretrained models often lack a dense neural thicket in their local weight-space neighborhood, and
eliciting task expertise appears to require more than random search. Characterizing when thickets
emerge, and when task expertise follows readily from pretraining, remains an open question.

%% file: statements.tex
\section*{AI use statement}

AI tools, including OpenAI's Codex and Anthropic's Claude Code, were used to assist with
experimentation and analysis code, figure generation, literature review, and manuscript editing.
Portions of the appendix were also drafted with AI assistance. AI assisted in analyzing the selection
noise null model, though it was suggested by the authors. The authors assume responsibility for
all content and claims in the paper.

%

\section*{Acknowledgments}
JA was supported by the National Science Foundation Graduate Research
Fellowship under Grant No.~DGE-1842487 and by a National Science Foundation award CCF-2239265.
GH acknowledges funding support from Coefficient Giving. We also acknowledge computational support
from UKRI and Isambard-AI.
VS was supported by National Science Foundation award 2239265 and 2346058, an Amazon Research Award,
a Google Research Scholar Award, and an Okawa Foundation Research Grant.

%% file: appendix.tex
\input{sections/related_work}

\section{Experimental Details}
\label{app:protocol}

\subsection{Computational details}

\textbf{Software.} Our implementation builds directly upon commit \texttt{536df0} of the official
\texttt{sunrainyg/RandOpt} repository of \citet{gan2026neuralthickets}. Unless explicitly noted
(e.g., in Appendix~\ref{app:implementation}), all our experiments use this code to generate weight perturbations. However, we deviate from the official
repository in one important respect. After evaluating a perturbed model, we restore the anchor weights exactly to their original values. This differs from the official implementation, which subtracts the perturbation in bf16 arithmetic, leaving a small rounding residual that can accumulate across candidates. We ablate this difference in Appendix~\ref{app:implementation} and find it to have little effect.
Code and compact artifacts supporting every figure and table are available at
\href{https://github.com/asil012/RevisitingNeuralThickets}
{\texttt{github.com/asil012/RevisitingNeuralThickets}}.

\textbf{Inference.} We use greedy decoding for all experiments by default, in keeping with Gan \&
Isola. When stochastic decoding is used, we sample with temperature $0.6$ and top-$p=0.95$
\citep{holtzman2020degeneration}. Countdown and GSM8K use a 1,024-token limit, while MBPP uses a
2,048-token limit. We use vLLM \citep{kwon2023vllm} with version 0.16.0 and PyTorch 2.9.1+cu128.

\textbf{RandOpt.} In keeping with the official implementation of Gan \& Isola,
for the $i$th
perturbation of the anchor model $\theta$, we draw $\sigma_i$ i.i.d.\ from the uniform distribution
over $\{0.5,1,2\}\times10^{-3}$ and then draw $\epsilon_i$ i.i.d.\ from the standard multivariate
Gaussian distribution.\footnote{We note that this differs slightly from the $\sigma$ range listed
in \citet{gan2026neuralthickets}, which instead uses $\{1, 2, 3\} \times 10^{-3}$. We follow
the released implementation, which uses the $\{0.5, 1, 2\} \times 10^{-3}$ range.}
The perturbed model is then $\theta_i=\theta+\sigma_i\epsilon_i$. We evaluate $N=5{,}000$ candidates for OLMo-3-7B-Instruct and Qwen2.5-3B/7B-Instruct, and a smaller
$N=1{,}000$ candidates for Qwen2.5-32B-Instruct. The top-$K$ candidates are selected by their mean
reward on the selection set. At test time, RandOpt aggregates the top-$K$ candidates by majority
vote over valid answers.

\textbf{Data.} By default, we use the first $m=200$ questions as the selection set for Countdown,
GSM8K, and MBPP. For Countdown, we use the following $1{,}400$ questions as the held-out test set.
For GSM8K and MBPP, we use the official test sets of $1{,}319$ and $500$ questions, respectively.
The separate Figure~\ref{fig:headline} (left) diagnostic uses
1,000 perturbations at $\sigma=10^{-3}$, splitting Countdown's test set into 700
selection questions and 700 held-out questions. The top-right panel uses another set
of 1,000 perturbations and samples random $m$-subsets from the $1{,}600$-question Countdown
pool.

\subsection{Scoring and voting}

We preserve the scoring rules of Gan \& Isola, which combine accuracy and
formatting for Countdown, and use binary task rewards for GSM8K and MBPP. For Countdown, the
selection reward decomposes as a sum of answer correctness $r_{\mathrm{ans}}$ and a format score
$r_{\mathrm{fmt}}$. In particular, the selection reward is
$r_{\mathrm{sel}}=r_{\mathrm{ans}}+0.1r_{\mathrm{fmt}}$. We often report test accuracy, which is
simply $r_{\mathrm{ans}}$. GSM8K simply records correctness of a numerical answer, while MBPP
requires the produced code solution to pass all supplied tests. In all cases, malformed and unparsed
outputs are counted as failures. We examine scoring sensitivity in Appendix~\ref{app:scorer-sensitivity}.
For voting, we use the same procedure for both RandOpt and TT-MV of the anchor model.
In particular, we exclude invalid answers before majority voting, and break ties in favor of the
earliest voter (i.e., higher selection rank for RandOpt).

\subsection{Uncertainty and density estimation}
\label{app:uncertainty}
\label{app:density-protocol}

\textbf{Uncertainty.} Our selection-subset intervals report the 2.5th and 97.5th
percentiles over 200 uniformly sampled selection sets, drawn without replacement while
holding candidate outputs fixed. (Thus, the full-pool endpoint has only one distinct subset
and lacks such intervals.) Paired-question 95\% bootstrap intervals use 2,000 joint resamples of
expert and anchor outcomes \citep{dror2018hitchhiker}, conditional on the recorded candidates
and voter orderings. Repeated answers remain grouped by question.

\textbf{Accuracy density.} Figures~\ref{fig:accuracy-density-m}
and~\ref{fig:accuracy-density-size} compare $\sigma=0.001$ perturbations with
greedy decoding (Table~\ref{tab:density-populations}) against 100 sampled anchor answer
sets at $T=0.6$, top-$p=0.95$.
Density is the fraction exceeding the specified multiple of each population's own mean
accuracy on the same questions, with half credit for ties (tolerance $10^{-12}$).
We average over 200 matched random question orderings, taking prefixes of length $m$
from the primary test sets; the model-size comparison fixes $m=200$.

\begin{table}[h]
\centering
\small
\begin{tabular}{lrrr}
\toprule
Model & Countdown & GSM8K & MBPP \\
\midrule
OLMo-3-7B-Instruct & 338 & 1,717 & 1,717 \\
Qwen2.5-0.5B-Instruct & 100 & 100 & 100 \\
Qwen2.5-1.5B-Instruct & 100 & 100 & 100 \\
Qwen2.5-3B-Instruct & 1,717 & 1,717 & 1,717 \\
Qwen2.5-7B-Instruct & 1,717 & 1,717 & 1,717 \\
Qwen2.5-32B-Instruct & 100 & 338 & 338 \\
\bottomrule
\end{tabular}
\caption{Unselected perturbation counts in the fixed-scale accuracy-density analysis
    (Figures~\ref{fig:accuracy-density-m} and~\ref{fig:accuracy-density-size}).
    Each setting uses 100 stochastic anchor rollouts.}
\label{tab:density-populations}
\end{table}

\section{The Selection-Noise Null Model}
\label{app:selection}

Under the selection-noise null model we consider, candidates share a reward distribution on each
question and differ only through independent evaluation noise. (Intuitively, it is as if each
candidate were simply a single rollout of a stochastic anchor model.) More precisely, for
candidate $i$ and question $j$, the model assumes that $Y_{ij}$, the candidate's reward
on that question, takes the form
\[  Y_{ij}=\mu_j+\varepsilon_{ij}. \]
Here $\mu_j$ is the common expected reward on question $j$, and $\varepsilon_{ij}$ is a zero-mean
random variable with variance $v_j$. The model assumes the errors $\varepsilon_{ij}$ to be
independent across candidate--question pairs.

Averaging $m$ questions gives score variance $v/m$, where
$v=m^{-1}\sum_{j\in S_m}v_j$.
Consequently, the Gaussian approximation in Section~\ref{sec:null} predicts
$\E[\Delta_{\mathrm{sel}}(m)]\approx a_N\sqrt{v/m}$. Note that the search size fixes
$a_N$; for our experiments with $N = 1{,}000$ and $N = 5{,}000$, we have $a_{1000}\approx3.241$
and $a_{5000}\approx3.678$. However, $v$ is data-dependent and must be estimated from the observed candidate rewards. (It is the only parameter of the Gaussian approximation that must be estimated.)

We estimate $v$ directly from the candidate rewards. For a question pool of size $Q$, one simply has
\begin{equation}
    \widehat\mu_j=\frac{1}{N}\sum_{i=1}^N Y_{ij},
    \qquad
    \widehat v=\frac{1}{Q}\sum_{j=1}^Q
        \frac{1}{N}\sum_{i=1}^N(Y_{ij}-\widehat\mu_j)^2.
    \label{eq:estimated-null-variance}
\end{equation}
The divisor $N$ gives the variance of the empirical distribution used for resampling.
For uniformly sampled question subsets, this pool average is the expected within-subset
variance. Binary rewards give $\widehat v_j=\widehat p_j(1-\widehat p_j)$.
Countdown instead uses its recorded correctness-plus-format reward. We report the
resulting $\widehat v$ values in Table~\ref{tab:null-variance} for every model--task setting.
We note that the null curves in \cref{fig:headline} (top right) and
\cref{fig:selection-transfer} use 120 empirical-resampling simulations
per selection-set size.

\begin{table}[ht]
\centering
\small
\setlength{\tabcolsep}{6pt}
\begin{tabular}{llrrr}
\toprule
Model & Task / analysis & Candidates $N$ & Questions $Q$ & $\widehat v$ \\
\midrule
OLMo 7B & Countdown, headline sweep & 1,000 & 1,600 & 0.0915 \\
 & Countdown, transfer diagnostic & 1,000 & 1,000 & 0.0904 \\
 & GSM8K & 5,000 & 1,000 & 0.0276 \\
 & MBPP & 5,000 & 400 & 0.1004 \\
\midrule[0.3pt]
Qwen 3B & Countdown & 5,000 & 1,000 & 0.0677 \\
 & GSM8K & 5,000 & 1,000 & 0.0817 \\
 & MBPP & 5,000 & 400 & 0.1090 \\
\midrule[0.3pt]
Qwen 7B & Countdown & 5,000 & 1,000 & 0.1358 \\
 & GSM8K & 5,000 & 1,000 & 0.0427 \\
 & MBPP & 5,000 & 400 & 0.0683 \\
\midrule[0.3pt]
Qwen 32B & Countdown & 1,000 & 200 & 0.1275 \\
 & GSM8K & 1,000 & 1,000 & 0.0327 \\
 & MBPP & 1,000 & 400 & 0.0409 \\
\bottomrule
\end{tabular}
\caption{Mean within-question reward variance in every selection-noise comparison.
    Values use raw reward-squared units, before multiplying lifts by 100 to express points.
    All rows use the original three-scale candidate populations.
    Qwen-32B Countdown has a fixed 200-question comparison rather than a selection-size sweep.}
\label{tab:null-variance}
\end{table}

\section{Additional Experimental Results}
\label{app:results}

\subsection{Selection and transfer}
\label{app:transfer}
\label{app:selection-subsets}

Figure~\ref{fig:headline} (top right) averages 200 uniform question subsets drawn without
replacement from the 1,600-question Countdown pool, for each $m\in\{25,50,100,200,400,800\}$. The full-pool endpoint uses the complete 1,600 questions once. We include
complementary OLMo selection-and-transfer sweeps in \cref{tab:decay-full},
comparing observed selection reward lift, the noise-null prediction, and held-out accuracy
gains. Selection lift falls as $m$ grows, while held-out gains over the perturbation
population remain near zero on GSM8K and modest on Countdown and MBPP.

\begin{table}[h]
\centering
\small
\setlength{\tabcolsep}{3.2pt}
\begin{tabular}{llrrr}
\toprule
Task & $m$ & Apparent lift & Noise null & Held-out rank-1 \\
\midrule
Countdown ($N=1{,}000$) & 25 & 18.40 & $18.03\pm3.07$ & $0.67\pm2.51$ \\
 & 50 & 13.37 & $12.89\pm1.66$ & $0.76\pm2.50$ \\
 & 100 & 10.08 & $9.38\pm1.09$ & $1.05\pm2.25$ \\
 & 200 & 7.69 & $6.73\pm0.68$ & $1.04\pm2.17$ \\
 & 400 & 5.98 & $4.78\pm0.52$ & $1.07\pm2.00$ \\
 & 800 & 4.80 & $3.42\pm0.34$ & $0.65\pm1.51$ \\
\midrule
GSM8K ($N=5{,}000$) & 25 & 9.63 & $8.57\pm3.34$ & $-0.20\pm1.33$ \\
 & 50 & 7.60 & $7.61\pm1.71$ & $-0.06\pm1.25$ \\
 & 100 & 5.90 & $5.68\pm0.95$ & $0.00\pm1.21$ \\
 & 200 & 4.58 & $4.06\pm0.50$ & $0.09\pm1.21$ \\
 & 400 & 3.50 & $2.96\pm0.32$ & $0.15\pm1.21$ \\
 & 800 & 2.72 & $2.10\pm0.16$ & $0.48\pm0.80$ \\
\midrule
MBPP ($N=5{,}000$) & 25 & 20.56 & $20.77\pm3.28$ & $0.43\pm3.10$ \\
 & 50 & 15.59 & $15.32\pm1.62$ & $0.94\pm2.80$ \\
 & 100 & 11.86 & $11.06\pm0.94$ & $1.18\pm2.80$ \\
 & 200 & 8.75 & $8.03\pm0.77$ & $1.30\pm2.80$ \\
 & 400 & 7.30 & $5.68\pm0.46$ & $1.72$ \\
\bottomrule
\end{tabular}
\caption{OLMo selection-and-transfer sweeps. Lift and null entries are reward points above the
    perturbation population; null entries give mean $\pm$ simulation SD.
    Held-out entries are accuracy points with the symmetrized
    95\% half-width of the selection-subset percentile interval shown after $\pm$.}
\label{tab:decay-full}
\end{table}

The Qwen sweeps show task-dependent decay: rank-1 lift falls steadily on
GSM8K and MBPP. On Countdown, it falls through $m=200$, then remains nearly flat
for Qwen-3B and declines more gradually for Qwen-7B
(Table~\ref{tab:qwen-selection-decay}).

\begin{table}[t]
\centering
\small
\begin{tabular}{llrrrr}
\toprule
Model & Task & $m=25$ & $m=200$ & $m=400$ & $m=800$ \\
\midrule
Qwen 3B & GSM8K & 14.97 & 9.11 & 7.77 & 6.87 \\
 & MBPP & 21.74 & 10.12 & 7.92 & --- \\
 & Countdown & 31.42 & 24.08 & 23.89 & 23.83 \\
\midrule[0.3pt]
Qwen 7B & GSM8K & 9.04 & 5.65 & 4.75 & 3.98 \\
 & MBPP & 16.65 & 6.87 & 5.51 & --- \\
 & Countdown & 31.66 & 18.16 & 16.76 & 16.14 \\
\midrule[0.3pt]
Qwen 32B & GSM8K & 7.01 & 4.31 & 3.51 & 3.00 \\
 & MBPP & 9.44 & 4.38 & 3.37 & --- \\
\bottomrule
\end{tabular}
\caption{Qwen2.5-Instruct rank-1 selection reward lift over the perturbation population
    (points) as selection-set size $m$ increases. Searches use $N=5,000$ candidates for 3B/7B
    and $N=1,000$ for 32B. Entries average 200 uniform question subsets drawn without replacement
    from 1,000 Countdown or GSM8K
    questions, or 400 MBPP tasks; the MBPP $m=400$ endpoint uses the full pool
    once (and $m=800$ is unavailable).}
\label{tab:qwen-selection-decay}
\end{table}

\textbf{Sensitivity to the selection questions.}
Naturally, individual OLMo experts are more sensitive to the selection questions than group averages.
At $m=200$ on Countdown, the selection-subset interval for held-out accuracy gain over the
perturbation population is $[-0.92,3.42]$ points for rank 1, versus $[0.51,1.49]$ for the
top-20 mean, despite average gains of about one point in both cases. The top-50 intervals
remain close to zero on GSM8K ($[-0.20,0.19]$) and slightly positive on MBPP ($[0.83,1.58]$).

Qwen-3B Countdown retains a larger advantage: increasing
$m$ from 200 to 800 barely changes the observed selection lift ($24.08$ to $23.83$ points),
while the null falls from $7.41\pm0.81$ to $3.56\pm0.32$
(Table~\ref{tab:qwen-selection-decay}).
At $m=200$, its top-50 held-out gain is $4.98$ accuracy points above the perturbation
population, with a selection-subset interval of $[4.33,5.63]$
(Table~\ref{tab:selection-robustness}). This strongly suggests
greater persistence than the
selection-noise null predicts among top candidates.

\begin{table}[H]
\centering
\small
\setlength{\tabcolsep}{4pt}
\begin{tabular}{llrrrr}
\toprule
Model & Task & Lift & Noise null & Null / lift & Top-50 transfer \\
\midrule
OLMo 7B & Countdown & 7.69 & 6.73 & $87.6\%$ & 0.93 \\
 & GSM8K & 4.58 & 4.06 & $88.6\%$ & 0.00 \\
 & MBPP & 8.75 & 8.03 & $91.8\%$ & 1.18 \\
\midrule[0.3pt]
Qwen 3B & Countdown & 24.08 & 7.41 & $30.8\%$ & 4.98 \\
 & GSM8K & 9.11 & 6.90 & $75.8\%$ & 3.90 \\
 & MBPP & 10.12 & 8.38 & $82.7\%$ & 2.83 \\
\midrule[0.3pt]
Qwen 7B & Countdown & 18.16 & 9.55 & $52.6\%$ & 6.40 \\
 & GSM8K & 5.65 & 4.85 & $85.9\%$ & 2.07 \\
 & MBPP & 6.87 & 6.69 & $97.5\%$ & 0.86 \\
\midrule[0.3pt]
Qwen 32B & Countdown & 8.55 & 7.96 & $93.1\%$ & --- \\
 & GSM8K & 4.31 & 3.66 & $85.1\%$ & 1.08 \\
 & MBPP & 4.38 & 4.38 & $100.1\%$ & 0.25 \\
\bottomrule
\end{tabular}
\caption{Selection-noise comparisons at $m=200$. Lift and null use reward points; top-50
    transfer averages individual held-out accuracy gains over the perturbation population.
    Entries average 200 selection subsets, except Qwen-32B Countdown, which uses the fixed
    first 200 questions and lacks a population-wide holdout comparison.
    Countdown transfer uses the auxiliary 600-question holdout; other tasks use their primary tests.
    OLMo Countdown and Qwen-32B use $N=1{,}000$; other settings use
    $N=5{,}000$.}
\label{tab:selection-robustness}
\end{table}

\subsection{Individual experts and anchor variability}
\label{app:single}
\label{app:scatter}
\label{app:fixed-repeats}

\textbf{Individual-expert comparisons.} Figure~\ref{fig:single} compares the selected experts
with five-run anchor means (Table~\ref{tab:fresh-anchors}), retaining their original ranks
and test outputs.
Table~\ref{tab:k1-full} preserves the earlier matched comparisons and
their paired-question intervals.

\begin{table}[h]
\centering
\begingroup
\small
\setlength{\tabcolsep}{4pt}
\begin{tabular}{llrrr}
\toprule
Model & Task & $N$ & Rank 1 / anchor & Difference [95\% interval] \\
\midrule
OLMo 7B & GSM8K & 5,000 & $85.44/87.41$ & $-1.97\;[-3.15,-0.76]$ \\
OLMo 7B & MBPP & 5,000 & $63.80/61.60$ & $+2.20\;[-1.40,5.80]$ \\
Qwen 3B & GSM8K & 5,000 & $81.12/80.67$ & $+0.45\;[-1.52,2.28]$ \\
Qwen 3B & MBPP & 5,000 & $56.00/56.60$ & $-0.60\;[-4.01,3.40]$ \\
Qwen 3B & Countdown & 5,000 & $26.79/15.43$ & $+11.36\;[8.93,14.07]$ \\
Qwen 7B & GSM8K & 5,000 & $87.95/88.02$ & $-0.08\;[-1.36,1.14]$ \\
Qwen 7B & MBPP & 5,000 & $63.20/65.00$ & $-1.80\;[-4.80,1.20]$ \\
Qwen 7B & Countdown & 5,000 & $43.71/33.36$ & $+10.36\;[7.64,13.07]$ \\
Qwen 32B & GSM8K & 1,000 & $91.51/90.07$ & $+1.44\;[0.15,2.73]$ \\
Qwen 32B & MBPP & 1,000 & $77.60/79.90$ & $-2.30\;[-5.10,0.30]$ \\
Qwen 32B & Countdown & 1,000 & $48.21/47.29$ & $+0.93\;[-1.43,3.14]$ \\
\bottomrule
\end{tabular}
\endgroup
\caption{Historical matched rank-1 solo performance on the same questions as the anchor. Countdown uses the common
    1,400-question test and MBPP the official 500-question test. The anchor column averages
    the recorded executions within each comparison. Differences are accuracy points, with
    paired-question 95\% bootstrap intervals conditional on the saved executions.
    Figure~\ref{fig:single} instead uses the five-run means in Table~\ref{tab:fresh-anchors}.}
\label{tab:k1-full}
\end{table}

In these historical matched comparisons, Qwen-3B and Qwen-7B show the
largest rank-1 gaps on Countdown. Outside Countdown, the only intervals excluding
zero are OLMo on GSM8K ($-1.97$ points) and Qwen-32B on GSM8K ($+1.44$ points).

\textbf{Selection and holdout.} Figure~\ref{fig:headline} (left) uses 1,000 fresh
$\sigma=0.001$ perturbations and a separate 700/700 split drawn uniformly at random
(seed 42), with selection--holdout
accuracy correlation $r=0.24$. Its rank-1 expert scores $69.86\%$ on selection and $69.00\%$
on holdout; the top-20 holdout mean is $68.68\%$.
The five-run anchor averages $69.06\%$ on holdout (range $68.43$--$69.71\%$), compared
with $66.74\%$ on selection. Thus the equally sized halves have different baseline
accuracies. The markers follow the 700-question selection ranking.

\textbf{Repeated execution of one expert.} To measure variability with fixed weights,
we reconstruct candidate 930 at $\sigma=5\times10^{-4}$ from a clean bf16 anchor and
evaluate both conditions greedily five times with generation seeds 42--46 on the same 1,400
Countdown questions. The expert averages $0.20$ points below the anchor, with a 95\% interval
of $[-1.63,1.23]$ (Table~\ref{tab:fixed-repeats}).
We obtain this interval by averaging the five paired differences within each question,
then bootstrapping questions 5,000 times.

\begin{table}[h]
\centering
\small
\begingroup
\begin{tabular}{rrrr}
\toprule
Generation seed & Anchor & Candidate 930 & Difference \\
\midrule
42 & 68.07 & 67.93 & $-0.14$ \\
43 & 67.86 & 67.79 & $-0.07$ \\
44 & 68.14 & 67.64 & $-0.50$ \\
45 & 68.07 & 67.57 & $-0.50$ \\
46 & 67.36 & 67.57 & $+0.21$ \\
\midrule
Mean & 67.90 & 67.70 & $-0.20$ \\
\bottomrule
\end{tabular}
\endgroup
\caption{Matched OLMo Countdown repeats on the same 1,400 questions. Accuracy is in percent;
    differences are candidate 930 minus anchor, in points.}
\label{tab:fixed-repeats}
\end{table}

Across the ten pairs of executions, mean final-answer disagreement is $6.06\%$ for the
anchor and $6.00\%$ for the expert; correctness disagreement is $5.81\%$ and $5.89\%$.
These rates compare extracted numerical answers, treating invalid responses as empty,
rather than comparing the full generated text.

\subsection{Validity and correctness on Countdown}
\label{app:countdown-decomposition}

This decomposition uses Qwen-3B, 7B, and 32B on the primary 1,400-question test, with both candidate and anchor
outcomes aligned to source rows 200--1599.
The marginal decomposition below retains the historical anchor executions.
The shared-valid-answer analysis in Appendix~\ref{app:common-valid} instead uses all five
fresh anchor runs, matching the updated comparisons in Section~\ref{sec:single}.

Countdown answers can fail because no valid equation is extracted or because a valid equation
misses the target. To distinguish these sources of improvement across the three Qwen sizes, write accuracy as
$c=vg$, where $v$ is the probability of an extractable valid equation and $g$ is correctness
conditional on validity. Subtracting the anchor's accuracy gives
\begin{equation}
    \Delta c=(\Delta v)g_0+v_0(\Delta g)+(\Delta v)(\Delta g).
\end{equation}
For a group of experts, we pool their candidate--question outcomes before taking
$g=\bar c/\bar v$; this is not the unweighted mean of each expert's conditional success rate.
The decomposition concerns binary test accuracy, without the $0.1r_{\mathrm{fmt}}$ selection
bonus. The validity flag checks extraction and the scorer's expression requirements, including
use of the supplied numbers. Without the underlying rejected responses, it cannot distinguish
tagging failures from every other reason an equation was rejected.

Table~\ref{tab:validity} reports the resulting decomposition across model sizes
and selection ranks.

\begin{table}[h]
\centering
\begingroup
\small
\setlength{\tabcolsep}{4pt}
\begin{tabular}{lrrrr}
\toprule
Group & Total gain & Validity & Conditional correctness & Interaction \\
\midrule
\multicolumn{5}{l}{Qwen-3B ($N=5{,}000$)} \\
Population & $-2.01$ & $-0.15$ & $-1.88$ & $+0.02$ \\
Top 20 & $+4.32$ & $+4.15$ & $+0.13$ & $+0.04$ \\
Top 50 & $+3.32$ & $+3.79$ & $-0.37$ & $-0.09$ \\
Selected rank 1 & $+11.36$ & $+6.91$ & $+3.07$ & $+1.38$ \\
\midrule[0.3pt]
\multicolumn{5}{l}{Qwen-7B ($N=5{,}000$)} \\
Population & $-1.00$ & $-1.53$ & $+0.56$ & $-0.03$ \\
Top 20 & $+6.18$ & $+4.08$ & $+1.87$ & $+0.23$ \\
Top 50 & $+5.23$ & $+3.25$ & $+1.80$ & $+0.18$ \\
Selected rank 1 & $+10.36$ & $+3.57$ & $+6.13$ & $+0.66$ \\
\midrule[0.3pt]
\multicolumn{5}{l}{Qwen-32B ($N=1{,}000$; selected candidates only)} \\
Top 20 & $-0.05$ & $-0.59$ & $+0.55$ & $-0.01$ \\
Top 50 & $-0.56$ & $-1.08$ & $+0.53$ & $-0.01$ \\
Selected rank 1 & $+0.93$ & $-2.04$ & $+3.10$ & $-0.13$ \\
\bottomrule
\end{tabular}
\endgroup
\caption{Qwen Countdown primary 1,400-question test gain decomposition,
    in accuracy points. Top-$K$ rows average individual experts, not their majority vote.}
\label{tab:validity}
\end{table}

Qwen-3B's anchor has $v_0=26.14\%$ and $g_0=59.02\%$. For its top 20,
$v=33.18\%$ and $g=59.52\%$: the baseline-weighted validity component is $4.15$ points,
approximately $96\%$ of the total gain. For the top 50, validity rises to $32.56\%$ while
conditional correctness falls to $57.60\%$; the $3.79$-point validity contribution exceeds
the $3.32$-point net gain because the other components are negative. Rank 1 has a different
balance: $v=37.86\%$ and $g=70.75\%$, so validity contributes about $61\%$ of its gain,
with positive conditional-correctness and interaction terms.

The Qwen-7B anchor has $v_0=48.71\%$ and $g_0=68.48\%$. Among the top 50, validity rises to $53.46\%$ and
conditional correctness to $72.18\%$, so most of the group gain comes from producing more valid
answers. For rank 1, validity reaches $53.93\%$ and conditional correctness $81.06\%$; improved
correctness accounts for the larger share of this candidate's gain. Changing which responses are valid also prevents a causal interpretation of the
conditional-correctness component as arithmetic ability alone.
At 32B, the anchor has $v_0=56.43\%$ and $g_0=83.80\%$. Its rank-1 expert has
lower validity ($54.00\%$) and higher conditional correctness ($89.29\%$), leaving a small
net gain. The top 20 and top 50 average $0.05$ and $0.56$ points below the anchor.

\subsubsection{Comparisons on the same valid-answer questions}
\label{app:common-valid}

The marginal decomposition above compares conditional accuracies on different
sets of valid responses. To compare the rank-1 expert and anchor on the same questions,
let $B$ denote questions where both produce valid expressions, $E$ those where only the
expert does, and $A$ those where only the anchor does.
Their total excess number of correct answers is exactly
\begin{equation}
    n\Delta c
      = \sum_{j\in B}(c_{ij}-c_{0j})
        + \sum_{j\in E}c_{ij}
        - \sum_{j\in A}c_{0j}.
    \label{eq:common-valid-gain}
\end{equation}
The first term compares correctness on shared valid-answer questions.
The other terms count correct answers gained and lost when the set of valid responses changes.
All three terms use the original primary test and the original selection ranking.
We compute the partition separately against each of the five fresh anchor executions
and average its components. The resulting total equals the expert's accuracy minus
the five-run anchor mean used in Figure~\ref{fig:single}.

Table~\ref{tab:common-valid} reports the resulting shared- and
changed-valid-set contributions.

\begin{table}[H]
\centering
\small
\setlength{\tabcolsep}{5pt}
\begin{tabular}{lrrrrr}
\toprule
Model & Both valid & \shortstack{Expert\\wins / losses} & \shortstack{Shared-set\\gain} & \shortstack{Valid-set\\change} & \shortstack{Per-run\\$p$ range} \\
\midrule
Qwen 3B & 177.8 & 31.0 / 20.6 & $+0.74$ & $+11.41$ & 0.025--0.401 \\
Qwen 7B & 481.4 & 61.0 / 34.2 & $+1.91$ & $+9.50$ & 0.002--0.045 \\
Qwen 32B & 571.4 & 25.0 / 12.0 & $+0.93$ & $+0.81$ & 0.024--0.090 \\
\bottomrule
\end{tabular}
\caption{Rank-1 Countdown gain split by shared versus changed valid-answer sets.
    Counts and gains average comparisons against five fresh greedy anchor runs.
    Both gain columns use accuracy points over all 1,400 questions and sum, before rounding,
    to the expert--anchor gap. Wins/losses count discordant correctness on shared valid-answer
    questions. The final column gives the range of unadjusted two-sided exact McNemar
    $p$-values computed separately for each run.}
\label{tab:common-valid}
\end{table}

Averaging over the five anchor comparisons, Qwen-3B produces valid expressions on
$352.2$ questions where the anchor does not, answering $233.6$ correctly, and loses validity
on $171.6$ questions, $73.8$ of which the anchor answered correctly.
These changes contribute $100(233.6-73.8)/1400=11.41$ accuracy points,
or $93.9\%$ of its $12.16$-point gain.
The shared set contributes the remaining $0.74$ points ($31.0$ wins, $20.6$ losses).
For Qwen-7B, changing the valid-answer set contributes $9.50$ points, or $83.2\%$ of its
$11.41$-point gain; shared-set correctness contributes $1.91$ points.
Qwen-32B gets $0.93$ of its $1.74$-point gain on the shared set.
Qwen-7B's shared-set advantage is significant at the unadjusted $5\%$ level against
each anchor execution; Qwen-3B's and Qwen-32B's are significant against one and three
of the five, respectively. These tests share the same expert and questions and do not
constitute five independent replications.
These are descriptive partitions of the observed gains: validity can fail for reasons
beyond formatting, and conditioning on both outputs being valid does not identify a
causal change in general reasoning ability.

\subsection{Alternative density definitions}
\label{app:density-variants}

\begin{figure}[H]
    \centering
    \includegraphics[width=\textwidth]{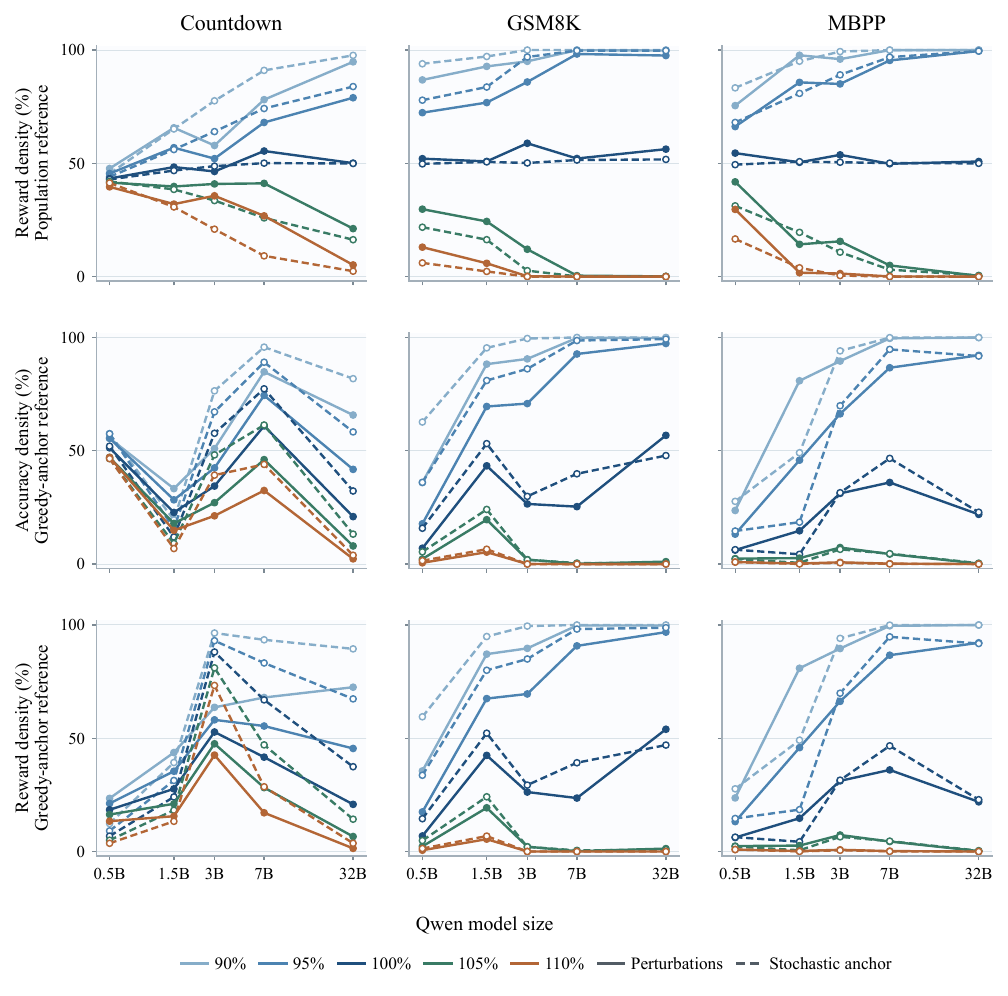}
    \caption{Density versus Qwen model size under alternative metrics and
        reference means ($m=200$). \textbf{Top:} Corrected task reward, with each
        family compared against its own population mean. \textbf{Middle:} Accuracy,
        using a greedy seed-42 anchor evaluation on the same question subset as the
        reference. \textbf{Bottom:} Corrected task reward using the same greedy-anchor
        reference. Solid curves show $\sigma=10^{-3}$ perturbations under greedy decoding;
        dashed curves show 100 stochastic anchor samples at $T=0.6$ and top-$p=0.95$, thresholded
        against the greedy anchor. Colors mark the threshold multiple of the reference
        mean, and ties receive half credit. Reward and accuracy coincide on MBPP.}
    \label{fig:density-size-variants}
\end{figure}

To test whether the model-size comparison depends on the metric or reference mean,
Figure~\ref{fig:density-size-variants} repeats Figure~\ref{fig:accuracy-density-size}
using corrected task reward and a greedy-anchor reference. The anchor-referenced variants
use one greedy seed-42 anchor evaluation on each matched question subset; both the
perturbation and stochastic-anchor populations are compared against this same reference.
Across the three tasks,
neither alternative produces a monotonic increase in favorable-perturbation density with
model size, while the stochastic-anchor curves retain qualitatively similar trends.

\clearpage

\raggedbottom
\section{Implementation Audit Details}
\label{app:implementation}

\subsection{Cross-tensor coupling}

The released sampler resets the random generator to the candidate's seed before perturbing
each tensor. As a result, tensors with the same shape and dtype receive identical noise
arrays. OLMo's 259 exposed tensors fall into six shape classes, so substantial sharing
occurs across layers, while each array still contains many random coordinates.
In our OLMo Countdown control, removing this coupling produces similar behavioral diversity:
the pairwise correctness-disagreement rates are $13.33\%$, $14.62\%$, and $16.86\%$
across the three perturbation scales with tied noise, versus $13.46\%$, $14.50\%$, and
$16.39\%$ with untied noise. Thus, this particular implementation choice has little effect
on the measured disagreement rates.

\subsection{Finite-precision restoration}
\label{app:restoration}

The reference implementation attempts to restore the anchor by subtracting each
perturbation after evaluation. Because bf16 addition and subtraction incur rounding,
$(w\oplus\sigma\epsilon)\ominus\sigma\epsilon$ can differ from $w$; successive candidates
can therefore accumulate weight drift. To assess its effect, we compare this procedure
with copying the stored anchor weights back after every candidate.

The two $N=5,000$ OLMo Countdown searches use the same candidate seeds, perturbation scales,
and worker assignments, with 1,250 candidates per worker across four inference workers.
The released and exact-restoration procedures select candidates 930 and 2948, respectively;
both answer 152 of the 200 selection questions correctly. These are distinct selected models
from a shared perturbation population. Table~\ref{tab:restoration-single} compares the
selection statistics and individual experts.

\begin{table}[h]
\centering
\small
\begin{tabular}{lrr}
\toprule
Quantity & Drift-on & Exact restore \\
\midrule
Selected candidate & 930 & 2948 \\
\midrule[0.3pt]
\multicolumn{3}{l}{\textit{Selection reward, $m=200$ (points)}} \\
Population mean & 74.027 & 73.984 \\
Across-candidate SD & 3.436 & 3.424 \\
Rank-1 reward & 82.840 & 82.715 \\
Top-50 mean lift over population & 6.549 & 6.428 \\
\midrule[0.3pt]
\multicolumn{3}{l}{\textit{Held-out accuracy (\%)}} \\
Rank-1 expert & 67.71 & 68.14 \\
Matched anchor & 67.14 & 68.14 \\
\midrule[0.3pt]
Rank-1 minus anchor (points, 95\% interval) & $+0.57\;[-1.36,2.43]$ & $0.00\;[-1.86,1.86]$ \\
\bottomrule
\end{tabular}
\caption{OLMo-3-7B-Instruct Countdown restoration control, with $N=5,000$ candidates per search.
    Selection statistics multiply raw rewards by 100; SD is across candidate mean rewards.
    Test accuracies use the same 1,400 questions. The final row gives paired-question 95\%
    bootstrap intervals conditional on the saved executions, using the original matched
    anchors rather than the later five-run baselines.}
\label{tab:restoration-single}
\end{table}

Despite the change in selected candidate, observed single-expert gains remain below one accuracy
point in both configurations. At the population level, exact restoration changes mean
selection reward by only $0.043$ points and the top-50 selection lift by $0.12$ points.
The held-out voting curves likewise remain close: their largest separation across the
evaluated budgets is $0.50$ accuracy points, falling to $0.07$ at $K=50$
(Table~\ref{tab:countdown-kcurve}).

\begin{table}[h]
\centering
\begingroup
\small
\begin{tabular}{rrrr}
\toprule
$K$ & Anchor TT-MV & RandOpt drift-on & RandOpt exact restore \\
\midrule
1 & 67.02 & 67.71 & 68.14 \\
2 & 74.84 & 74.57 & 74.57 \\
3 & 78.47 & 77.21 & 77.29 \\
5 & 81.81 & 80.14 & 80.29 \\
10 & 85.71 & 84.14 & 83.93 \\
20 & 89.02 & 87.79 & 87.86 \\
30 & 90.37 & 89.64 & 89.50 \\
40 & 91.25 & 90.50 & 90.43 \\
50 & 91.64 & 90.93 & 90.86 \\
\bottomrule
\end{tabular}
\endgroup
\caption{OLMo Countdown vote curves. Figure~\ref{fig:headline} (bottom right) shows the anchor and released
    restoration (drift-on) columns on the common 1,400-question test. Anchor values use the stored 50-sample bank
    during figure generation. The exact-restoration control uses the same candidate seed list.}
\label{tab:countdown-kcurve}
\end{table}

The
$K=1$ TT-MV entry averages the accuracies of the 50 individual anchor samples, giving
$67.02\%$. For $K=3$ through 40,
we average 20 random voter orderings with seed 20260908; at $K=50$, we use the entire bank in its
stored order. The added $K=2$ point uses 20 separately seeded orderings, preserving all existing
curve values. Voting retains the released valid-answer filtering and first-seen tie rule.
All curves are recomputed after restricting the saved outputs to the common 1,400 test questions;
no new model generations are involved. The full RandOpt curves
continue to $K=500$, but Wave 1 has only 50 anchor samples, so our matched comparisons stop at
$K=50$.

At $K=50$, the RandOpt-minus-anchor gaps are $-0.71\;[-1.79,0.36]$ points for the released
restoration arm and $-0.79\;[-1.71,0.21]$ for exact restoration. Both intervals include zero.

We also check whether candidate scores trend with the number of perturb--restore cycles,
as would be expected if accumulated drift systematically improved or degraded performance.
Regressing mean selection reward on cycle index separately for each worker gives slopes
between $-0.523$ and $+0.832$ reward points per 1,000 cycles with drift, compared with
$-0.190$ to $+0.516$ under exact restoration. Slopes vary in sign across workers in both
runs. Together with the closely matched population and voting results, this control provides
little evidence that restoration drift accounts for the missing large expert gain in this
OLMo Countdown setting.

\subsection{Decoder instability}
\label{app:anchor-repeats}

Even with greedy decoding, repeatability depends on the execution configuration.
Two anchor runs using batches of 200 produce identical final Countdown answers, despite
possible differences in the generated text. By contrast, two evaluations that each generated
2,000 answers differ on 140 of the primary test questions: both answer and correctness
disagreement reach $10.00\%$, while accuracy moves from $67.14\%$ to $68.14\%$.
We compute these comparisons from the saved outputs, preserving the original batching.

Repeating the same 1,000 exactly restored perturbations on 200 selection questions also
reveals variation: $12.34\%$ of answers and $12.30\%$ of correctness flags change.
However, those executions used different batch compositions, so their disagreement rates
cannot isolate a difference in reproducibility between experts and the anchor.
For that comparison, we use the matched fixed-candidate repeats in
Appendix~\ref{app:fixed-repeats}.

\textbf{Five-run anchor baselines.} For every model--task pair in Figure~\ref{fig:single},
we collect five greedy executions in fresh processes with seeds 42--46, fixed question
order and batching, and unchanged weights. Table~\ref{tab:fresh-anchors} reports their
means and observed variation. These runs use vLLM 0.16.0, PyTorch 2.9.1+cu128, and bf16;
OLMo and Qwen-3B/7B run on RTX 4090 GPUs with tensor-parallel size one, while Qwen-32B
uses two H100 GPUs. Eager execution is enabled and prefix caching disabled.
This bank supplies the Figure~\ref{fig:single} baselines and is separate from the
matched candidate-930 experiment in Appendix~\ref{app:fixed-repeats}.

\begin{table}[H]
\centering
\small
\setlength{\tabcolsep}{6pt}
\begin{tabular}{llrrrr}
\toprule
Model & Task & Mean & Minimum & Maximum & SD \\
\midrule
OLMo 7B & Countdown & 67.27 & 66.71 & 68.21 & 0.57 \\
 & GSM8K & 87.48 & 87.26 & 87.64 & 0.16 \\
 & MBPP & 61.72 & 60.40 & 62.60 & 0.88 \\
\midrule[0.3pt]
Qwen 3B & Countdown & 14.63 & 14.36 & 15.00 & 0.24 \\
 & GSM8K & 80.55 & 80.29 & 81.12 & 0.36 \\
 & MBPP & 55.48 & 55.40 & 55.80 & 0.18 \\
\midrule[0.3pt]
Qwen 7B & Countdown & 32.30 & 32.07 & 32.50 & 0.17 \\
 & GSM8K & 87.57 & 87.34 & 87.79 & 0.19 \\
 & MBPP & 65.08 & 64.80 & 65.60 & 0.33 \\
\midrule[0.3pt]
Qwen 32B & Countdown & 46.47 & 46.29 & 46.79 & 0.19 \\
 & GSM8K & 89.58 & 89.46 & 89.76 & 0.13 \\
 & MBPP & 80.36 & 80.20 & 80.40 & 0.09 \\
\bottomrule
\end{tabular}
\caption{Accuracy across five fresh greedy anchor executions, in percent.
    SD uses divisor four; the minimum and maximum describe the observed runs, not a
    confidence interval. Each execution evaluates 1,400 Countdown, 1,319 GSM8K,
    or 500 MBPP questions.}
\label{tab:fresh-anchors}
\end{table}

\subsection{Scoring sensitivity and execution variability}
\label{app:scorer-sensitivity}

\textbf{Sensitivity to answer formatting.} On GSM8K, even a trailing punctuation mark
can prevent a correct answer from receiving credit. We therefore rescore the saved answers
with a more tolerant comparator: after trimming whitespace and removing one trailing period
or comma, a leading dollar sign, and thousands separators, we compare numerically whenever
the prediction and reference both parse as numbers. Throughout this check, the extracted
answers, expert rankings, winning vote identities, and original paired anchor executions
remain fixed. Only the answer-acceptance rule changes.

\begin{table}[h]
\centering
\begingroup
\small
\setlength{\tabcolsep}{4pt}
\begin{tabular}{lrrrr}
\toprule
Model & Anchor accuracy & Rank-1 gap & Top-50 enrichment & Voting gap ($K=50$) \\
\midrule
OLMo 7B & $87.41/92.46$ & $-1.97/-2.24$ & $-0.05/-0.13$ & $+0.56/+0.25$ \\
Qwen 3B & $80.67/85.52$ & $+0.45/-0.61$ & $+3.91/+1.26$ & $-0.61/-0.47$ \\
Qwen 7B & $88.02/91.81$ & $-0.08/-0.08$ & $+2.20/+0.69$ & $-0.28/-0.64$ \\
Qwen 32B & $90.07/95.75$ & $+1.44/+0.61$ & $+1.18/+0.01$ & $+0.32/+0.11$ \\
\bottomrule
\end{tabular}
\endgroup
\caption{GSM8K reference/tolerant scoring on the same 1,319 test questions. Anchor accuracy
    is in percent; the other entries are accuracy-point differences. Rank-1 and voting gaps
    are relative to the anchor; top-50 enrichment is relative to the perturbation population.}
\label{tab:scorer-sensitivity}
\end{table}

As Table~\ref{tab:scorer-sensitivity} shows, this more lenient rule raises anchor accuracy
and reduces some apparent expert gains. In particular, Qwen-3B's rank-1 gap changes sign,
while Qwen-32B's top-50 enrichment falls from $1.18$ to $0.01$ points. The voting results,
however, remain close: across all four models, the RandOpt--anchor gap at $K=50$ stays below
one point in magnitude under either convention. We retain reference scoring for the primary
results; the alternative provides a sensitivity check.

Even within the reference implementation, selection reward and test accuracy need not
agree. For example, in one GSM8K anchor execution over 2,319 questions, 86 answers pass the
accuracy check yet receive zero selection reward. Countdown separates the two explicitly
through its $0.1$-weighted format bonus, which contributes approximately $0.83$ points to
the OLMo top-50 selection advantage.

Direct comparisons with the published results also depend on scorer version. After
publication, a correction replaced the stored anchor reward with correctness and revised
Countdown equation parsing. These changes further complicate reproduction of the original
evaluation pipeline.

\textbf{Execution variability in MBPP.} Grading MBPP requires running generated programs,
which may fail to terminate or allocate excessive memory. Execution safeguards are therefore
necessary, and we count failed generations in the accuracy denominator.

We also found conflicting grades for outputs sharing the same program hash. Inspection of
53 such pairs, spanning seven test questions and two selection questions, confirmed that
every pair contains byte-identical extracted code and correctly stored hashes. Thus hash
collisions do not explain these discrepancies. The failing outputs also carry no flag
indicating rejection by an execution safeguard.

For one affected program, changing Python's string-hash seed yields four passes in 20 clean
executions: scoring-side randomness can indeed produce conflicting verdicts for identical
code. The same test, however, did not reproduce the other conflicts: the checked
programs on the other eight questions remain stable across these hash seeds.
Worker state or execution timing may explain those discrepancies, but the available evidence
does not identify the cause.

Despite these inconsistencies, none of the identified conflicts changes a recorded winning
RandOpt or anchor vote at the evaluated budgets. We therefore retain the original grades.
Assessing their effect on selection ranks would require broader rescoring through the
production procedure.

%% file: sections/related_work.tex
\section{Related Work}
\label{sec:related}

\noindent\textbf{Neural thickets and structured expert search.}
The neural thicket was introduced by \citet{gan2026neuralthickets}, whose terminology and techniques
we adopt. In particular, \citet{gan2026neuralthickets} were the first to propose that pretrained weights
are surrounded by a dense population of task experts.
Subsequent work has examined whether random search can be replaced with a more efficient technique
or heuristic.
For instance, \citet{lzicar2026auditing} scores an initial batch of perturbations, uses those
scores to identify promising directions, and restricts further search to combinations of
such directions. However, their Qwen-7B Countdown audit finds no consistent advantage of learned
directions over their search controls. In a complementary approach, \citet{cendra2026bayesian} use
a probabilistic model of observed rewards to choose successive perturbations within a fixed set of random
directions. On small Qwen2.5-Instruct models, they find that
200 evaluations using their approach often match or exceed 1,000-draw RandOpt for single-expert
test accuracy. However, they also document cases where further selection gains yield flat or
even declining test accuracy.

\medskip
\noindent\textbf{Random search and low-dimensional adaptation.}
There is an ample body of work demonstrating that task adaptation of pretrained models often
requires only a low-dimensional parameter update. For instance, parameter-efficient fine-tuning
(PEFT) adapts a pretrained model by training a
small task-specific parameter set while freezing its core weights, as in inserted adapters
and learned soft prompts \citep{houlsby2019parameter,lester2021power}.
In particular, LoRA learns adapters as low-rank decompositions of weight updates, permitting
far fewer trainable parameters than the pretrained model \citep{hu2022lora}.
Subsequent work such as TinyLoRA, VeRA, and LoRA-XS extends this finding by designing adapters
with even fewer trainable parameters \citep{morris2026tiny,kopiczko2024vera,balazy2025loraxs}.
Most recently, \citet{lu2026deleting} reported successful task adaptation
by learning a binary mask that zeros a small fraction of pretrained weights while leaving
the retained values unchanged. Overall, such lightweight adaptations demonstrate the flexibility of
pretrained representations.

\medskip
\noindent\textbf{Ensembling and test-time computation.}
\citet{mehrtash2020pep} demonstrated the value of simple Gaussian weight
perturbations through Parameter Ensembling by Perturbation (PEP), which tunes the
perturbation variance of such Gaussian perturbations on validation data and then ensembles
the resulting predictions at test time. PEP serves as a direct precedent for the RandOpt
work of \citet{gan2026neuralthickets}, which adds the step of selecting individual perturbations
on task-specific data before aggregation.
More broadly, additional work has studied how a
single model benefits from greater test-time compute, including majority voting and
repeated sampling
\citep{wang2023selfconsistency,brown2024large,snell2025scaling,wu2025inference}.
Self-consistency, for example, samples multiple
reasoning paths and returns their modal answer \citep{wang2023selfconsistency}; at larger
budgets, repeated sampling can substantially increase the probability that at least one
output is correct \citep{brown2024large}. Interestingly, training method can also affect this coverage:
\citet{ba2026understanding} report higher pass@$K$ for models trained with evolution
strategies than with GRPO in their evaluated settings.